\documentclass[11pt]{article}
\usepackage{coling2014}
\usepackage{times}
\usepackage{url}
\usepackage{latexsym}
\usepackage{graphicx}
\usepackage{amsfonts, amssymb, amsmath}
\usepackage{algorithm, algorithmic, amsthm}
\usepackage{graphicx}
\usepackage{color}
\usepackage{hyperref}
\usepackage[top=2.5cm, bottom=3cm, left=3cm, right=3cm]{geometry}

\usepackage{subfigure}

\renewcommand{\b}{\ensuremath{\mathbf{b}}}

\newcommand{\e}{\ensuremath{\mathbf{e}}}

\renewcommand{\t}{\ensuremath{\mathbf{t}}}

\newcommand{\w}{\ensuremath{\mathbf{w}}}
\newcommand{\x}{\ensuremath{\mathbf{x}}}

\newcommand{\z}{\ensuremath{\mathbf{z}}}

\newcommand{\calS}{\ensuremath{\mathcal{S}}}

\newcommand{\calV}{\ensuremath{\mathcal{V}}}

{%
\begin{list}{#1}{
\vspace{-\topsep}
\vspace{-\partopsep}
\setlength{\itemindent}{0cm}
\setlength{\rightmargin}{0cm}
\setlength{\listparindent}{0cm}
\settowidth{\labelwidth}{#1}
\setlength{\leftmargin}{\labelwidth}
\addtolength{\leftmargin}{\labelsep}
\setlength{\itemsep}{0cm}
}%
}%
{%
\end{list}
\vspace{-\topsep}
\vspace{-\partopsep}
}

{\begin{enumerate}%
}%
{\end{enumerate}}

\title{$gen$CNN: A Convolutional Architecture for Word Sequence Prediction}

\author{Mingxuan Wang$^1$\ Zhengdong Lu$^2$ \ 
	Hang Li$^2$ \ Wenbin Jiang$^1$ \ Qun Liu$^{3,1}$ \\
	$^1$Institute of Computing Technology, Chinese Academy of Sciences\\
	{\tt \{wangmingxuan,jiangwenbin,liuqun\}@ict.ac.cn}\\
	$^2$Noah's Ark Lab, Huawei Technologies\\
	{\tt \{Lu.Zhengdong,HangLi.HL\}@huawei.com}\\
	$^3$Centre for Next Generation Localisation, Dublin City University\\
}

\begin{document}
\maketitle
\begin{abstract}
We propose a novel convolutional architecture, named $gen$CNN, for word sequence prediction. Different from previous work on neural network-based language modeling and generation (e.g., RNN or LSTM), we choose not to greedily summarize the history of words as a fixed length vector. Instead, we use a convolutional neural network to predict the next word with the history of words of variable length. Also different from the existing feedforward networks for language modeling, our model can effectively fuse the local correlation and global correlation in the word sequence,  with a convolution-gating strategy specifically designed for the task. We argue that our model can give adequate representation of the history, and therefore can naturally exploit both the short and long range dependencies. Our model is fast, easy to train, and readily parallelized. Our extensive experiments on text generation and $n$-best re-ranking in machine translation show that $gen$CNN outperforms the state-of-the-arts with big margins.
\end{abstract}

\section{Introduction}  
Both language modeling~\cite{wu2003maximum,rnnMikolov,Bengio03aneural} and text generation~\cite{Axelrod} boil down to modeling the conditional probability of a word given the proceeding words. Previously, it is mostly done through purely memory-based approaches, such as $n$-grams, which cannot deal with long sequences and has to use some heuristics (called smoothing) for rare ones. Another family of methods are based on distributed representations of words, which is usually tied with a  neural-network (NN) architecture for estimating the conditional probabilities of words.

Two categories of neural networks have been used for language modeling: 1) recurrent neural networks (RNN), and 2) feedfoward network (FFN): \vspace{-5pt}
\begin{itemize}
  \item The RNN-based models, including its variants like  LSTM, enjoy more popularity, mainly due to their flexible structures for processing word sequences of arbitrary lengths, and their recent empirical success\cite{sutskever2014sequence,SequenceGen_Graves13}. We however argue that RNNs, with their power built on the recursive use of a relatively simple computation units, are forced to make greedy summarization of the history and consequently not efficient on modeling word sequences, which clearly have a bottom-up structures. \vspace{-5pt}
  \item The FFN-based models, on the other hand,  avoid this difficulty by directly taking the history as input. However the FFNs are fully-connected networks, rendering them inefficient on capturing local structures of languages. Moreover their ``rigid" architectures make it futile to handle the great variety of patterns in long range correlations of words. \vspace{-5pt}
\end{itemize}

\begin{figure*}[t!]
\begin{center}
      \includegraphics[width=0.9\textwidth]{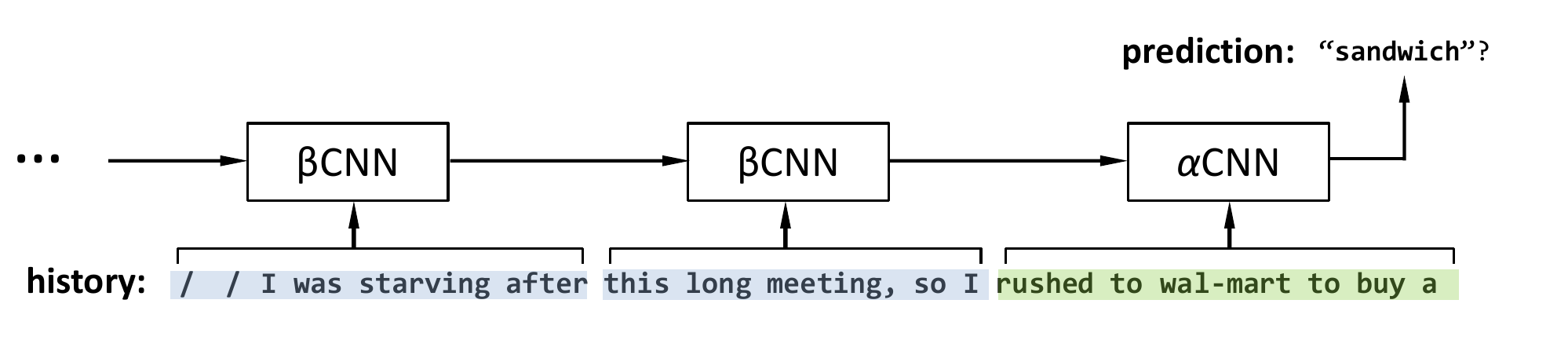}
 \caption{The overall diagram of a $gen$CNN. Here ``$/$" stands for a zero padding. In this example, each CNN component covers 6 words, while in practice the coverage is 30-40 words.}
    \label{f:overall}
  \end{center} \vspace{-10pt}
\end{figure*}

We propose a novel convolutional neural network architecture, named $gen$CNN, for efficiently combining local and long range structures of language with the purpose of modeling conditional probabilities.  $gen$CNN can be directly used in generating a word sequence (i.e., text generation) or evaluating the likelihood of a word sequence (i.e., language modeling). We also show the empirical superiority of $gen$CNN on both tasks over traditional $n$-grams and its RNN and FFN counterparts.

\paragraph{Notations:}
%In the remainder of the paper, we first give an overview of $gen$CNN (Section~\ref{s:overview}), and then elaborate on its architecture (Section~\ref{s:arch}) and learning (Section~\ref{s:learning}).  After that, we report our empirical study of $gen$CNN on text generation (Section~\ref{s:expts_generation}) and language modeling (Section~\ref{s:expts_LM}) In Section~\ref{s:related}, we wrap up the paper with a discussion on related work.
We use $\calV$ to denote the vocabulary, $\e_t$ ($\in \{1,\cdots, |\calV|\}$) to denote the $t^{th}$ word in a sequence $\e_{1:T} \overset{\text{def}}{=}  [\e_1,\cdots,\e_{T}]$, and $\e_t^{(n)}$ if the sequence itself is further indexed by $n$.

%\newpage

\section{Overview} \label{s:overview}
As shown in Figure~\ref{f:overall}, $gen$CNN is overall recursive, consisting of CNN-based processing units  of two types: 
\begin{itemize}
  \item $\alpha$CNN as the ``front-end", dealing with the history that is closest to the prediction; 
  \item $\beta$CNNs (which can repeat), in charge of more ``ancient" history. 

\end{itemize}
Together, $gen$CNN takes history $\e_{1:t} $ of arbitrary length to predict the next word $\e_{t+1}$ with probability 
\begin{equation}
p(\e_{t+1}\;|\e_{1:t};  \bar{\Theta}),
\label{e:condprob} 
\end{equation}
based on a representation $\phi(\e_{1:t}; \bar{\Theta})$ produced by the CNN, and a $|\calV|$-class soft-max: 
\begin{equation} \hspace{-6pt}
p(\e_{t+1}| \e_{1:t};  \bar{\Theta}) \propto e^{\mu_{\e_{t+1}}^\top \phi(\e_{1:t}) + b_{\e_{t+1}}}. 
\label{e:softmax}
\end{equation}
%A sentence can be generated by recursively sampling $\e_{t+1}$:  %\vspace{-5pt}
%$
%\e_{t+1}^{\star} \sim p(\e_{t+1}|\e_{1:t}; \bar{\Theta}). %\vspace{-5pt}
%$
$gen$CNN is fully tailored for modeling the sequential structure in natural language, notably different from conventional CNN~\cite{cnnCV,hu2014convolutional} in 1) its specifically designed weights-sharing strategy (in $\alpha$CNN), 2) its gating design, and 3) certainly its recursive architectures. Also distinct from RNN, $gen$CNN gains most of its processing power from the heavy-duty processing units (i.e.,$\alpha$CNN and $\beta$CNNs), which follow a bottom-up information flow and yet can adequately capture the temporal structure in word sequence with its convolutional-gating architecture.

%\begin{itemize}
%  \item specifically designed strategy for sharing convolution weights, enabling it to fuse the ``parsing" of the history and the prediction task of the prediction; \vspace{-4pt}
%  \item separate gating network instead of direct pooling over feature-maps for better composition;\vspace{-4pt}
%  \item recursive architecture to handle sequences of arbitrary length.
%\end{itemize}

\section{$gen$CNN: Architecture} \label{s:arch}
We start with discussing the convolutional architecture of $\alpha$CNN as a stand-alone sentence model, and then proceed to the recursive structure. After that we give a comparative analysis on the mechanism of $gen$CNN.

\begin{figure}[h!]
\begin{center}
    \hspace{-15pt}\includegraphics[width=0.7\textwidth]{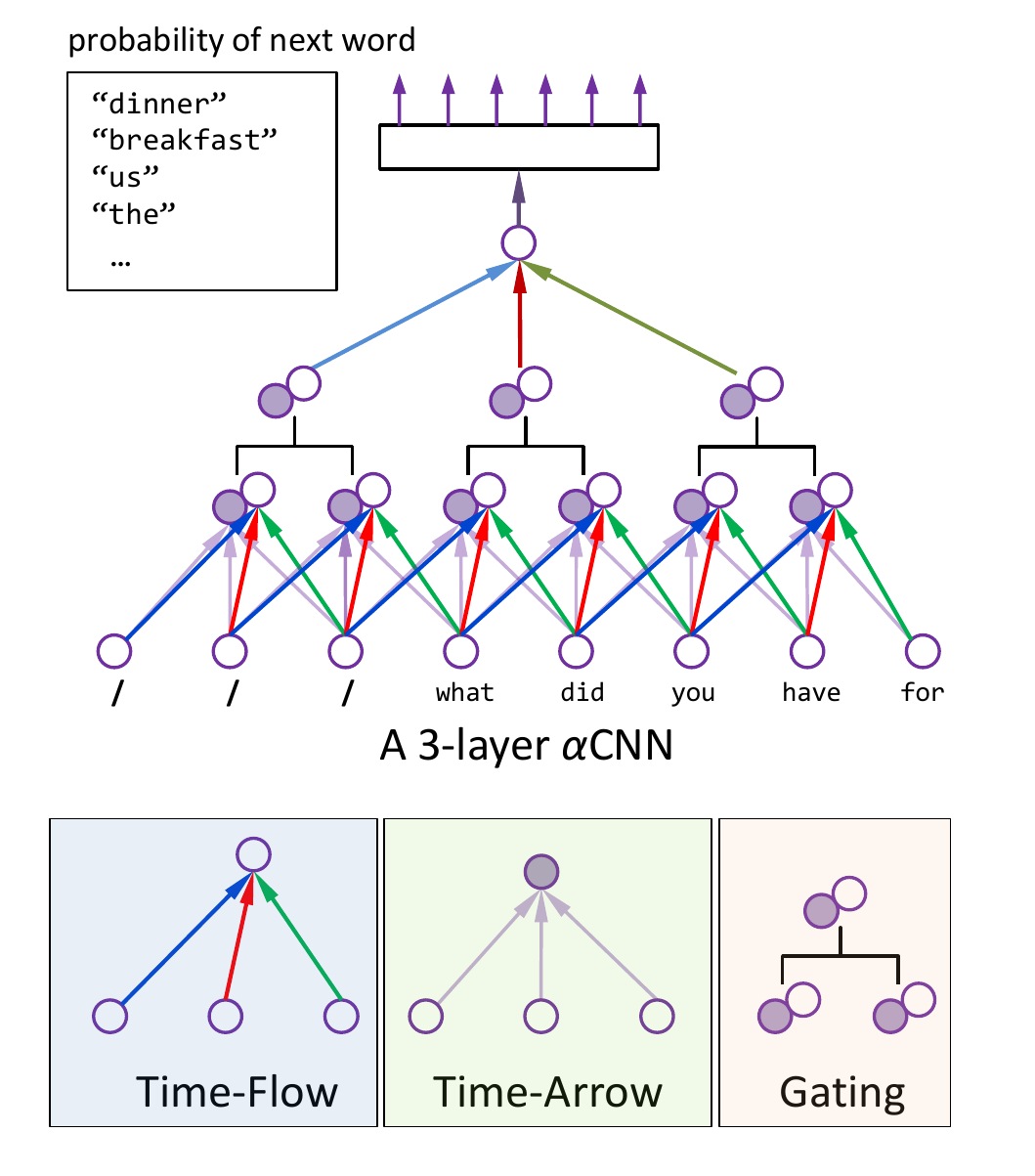} 
 \caption{Illustration of a 3-layer $\alpha$CNN. Here the unfilled nodes stand for the \textsc{Time-Time} feature-maps, and the 
 the filled nodes for \textsc{Time-Arrow}.}
    \label{f:skCNN}
  \end{center}
\end{figure}

$\alpha$CNN, just like a normal CNN, has fixed architecture with predefined maximum words (denoted as $L_\alpha$). History shorter than $L_\alpha$ will filled with zero paddings, and history longer than that will be folded to feed to $\beta$CNN after it, as will be elaborated in Section~\ref{s:recursive}. Similar to most other CNNs, $\alpha$CNN alternates between convolution layers and pooling layers, and finally a fully connected layer to reach the representation before soft-max, as illustrated by Figure~\ref{f:skCNN}. Unlike the toyish example in Figure~\ref{f:skCNN}, in practice we use a larger and deeper $\alpha$CNN with $L_\alpha= 30$ or $40$, and two or three convolution layers (see Section~\ref{s:details}). 

In Section~\ref{s:convolution} we will introduce the hybrid design of convolution in $gen$CNN for capturing structures of different nature in word sequence prediction. In Section~\ref{s:gate}, we will discuss the design of gating mechanism.

%
%Different from conventional CNN, $gen$CNN has 1) weight sharing strategy for convolution, and 2)``external" gating networks to replace the normal pooling mechanism, both of which are specifically designed for word sequence prediction.

%
%for history shorter than $L_{max}$, and recursive structure designed for handling history longer than $L_{max}$ (see Section~\ref{s:recursive}) . [need to rephrase]

%Figure \ref{f:1DCNNgen}
%\begin{figure}[h!]
%\begin{center}
%      \includegraphics[width=0.48\textwidth]{pics/CNNgen1D_v2.png}\\
%    \caption{Illustration for a 2-layer skewed-CNN as the front end of \text{$gen$CNN}.}
%    \label{f:1DCNNgen}
%  \end{center} %\vspace{-10pt}
%
%\end{figure}

%\begin{figure*}[t!]
%\begin{center}
%\includegraphics[width=0.48\textwidth]{pics/regularCNN.png} \hspace{10pt}
%      \includegraphics[width=0.48\textwidth]{pics/CNNgen1D_v4.png}\\ \
% \caption{$\alpha$CNN as the front end of $gen$CNN. Here the shadow...}
%    \label{f:1DCNNgen}
%  \end{center} %\vspace{-10pt}
%\end{figure*}
%

%\subsubsection{Convolution} %\vspace{-10pt} \label{s:convunit}

\subsection{$\alpha$CNN: Convolution}\label{s:convolution}
Different from conventional CNN, the weights of convolution units in $\alpha$CNN is only partially shared. More specifically, in the convolution units there are two types feature-maps: \textsc{Time-Flow} and the \textsc{Time-Arrow}, illustrated respectively with the unfilled nodes and filled nodes in  Figure~\ref{f:skCNN}. The parameters for \textsc{Time-Flow} are shared among different convolution units, while for \textsc{Time-Arrow} the parameters are location-dependent. Intuitively, \textsc{Time-Flow} acts more like a conventional CNN (e.g., that in~\cite{hu2014convolutional}), aiming to understand the overall temporal structure in the word sequences; \textsc{Time-Arrow}, on the other hand, works more like a traditional NN-based language model~\cite{nplm,Bengio03aneural}: with its location-dependent parameters, it focuses on capturing the prediction task and the direction of time.

For sentence input $\mathbf{x} \hspace{-3pt}=\hspace{-3pt} \{\x_1,\cdots, \x_T\}$, the feature-map of type-$f$ on Layer-$\ell$ is 

\noindent if $f \in $\textsc{ Time-Flow:}  
\begin{equation} %\small
%z^{(\ell,f)}_{i} \overset{\text{def}}{=}
z^{(\ell,f)}_{i}(\mathbf{x}) =  \sigma(\w^{(\ell,f)}_{\textsf{TF}} \hat{\mathbf{z}}^{(\ell-1)}_{i} + b_{\textsf{TF}}^{(\ell,f)}),
\end{equation}
\noindent if $f \in $\textsc{ Time-Arrow:}  
\begin{equation} %\small
%z^{(\ell,f)}_{i} \overset{\text{def}}{=}
z^{(\ell,f)}_{i}(\mathbf{x}) =  \sigma(\w^{(\ell,f,i)}_{\textsf{TA}} \hat{\mathbf{z}}^{(\ell-1)}_{i} + b^{(\ell,f,i)}_{\textsf{TA}}),
\end{equation}  
%As shown in Figure \ref{f:senCNN}, the convolution in Layer-1 operates on sliding windows of words (width $k_1$), and the similar definition of windows carries over to higher layers. Generally,with sentence input $\mathbf{x}$, the convolution unit for feature map of type-$f$ (among $F_\ell$ of them)  on Layer-$\ell$ is
%\begin{multline}
%z^{(\ell,f)}_{i} \overset{\text{def}}{=}
%z^{(\ell,f)}_{i}(\mathbf{x}) =  \sigma(\w^{(\ell,k)} \hat{\mathbf{z}}^{(\ell-1)}_{i} + b^{(\ell,f)}),\\ f = 1,2,\cdots, F_{\ell},
%\end{multline}
where 
\begin{itemize}
  \item $z^{(\ell,f)}_{i}(\mathbf{x})$ gives the output of feature-map of type-$f$  for location $i$ in Layer-$\ell$;

   \item $\sigma(\cdot)$ is the activation function, e.g., Sigmoid or Relu~\cite{relu} 
  \item $(\w^{(\ell, f)}_{\textsf{TF}}, \b^{(\ell, f)}_{\textsf{TF}})$ denotes the location-independent parameters for $f \hspace{-2pt}\in \hspace{-2pt}$ \textsc{Time-Flow} on Layer-$\ell$, while $(\w^{(\ell, f,i)}_{\textsf{TA}}, b^{(\ell, f,i)}_{\textsf{TA}})$ stands for that for $f \hspace{-2pt}\in\hspace{-2pt}$
   \textsc{Time-Arrow} and location $i$ on Layer-$\ell$;  
  \item $\hat{\mathbf{z}}^{(\ell-1)}_{i}$ denotes the segment of Layer-$\ell\hspace{-3pt}-\hspace{-3pt}1$ for the convolution at location $i$ , while
      \begin{equation*}
      \hat{\mathbf{z}}^{(0)}_{i} \overset{\text{def}}{=}  [ \x_{i}^\top, \;\, \x_{i+1}^\top,\;\cdots,\;\, \x_{i+k_1-1}^\top]^\top
      \end{equation*}
      concatenates the vectors for $k_{1}$ words from sentence input $\mathbf{x}$.
\end{itemize}

%\paragraph{Max-Pooling}%\vspace{-10pt}
%We take a max-pooling in every two-unit window for every $f$, after each convolution
%\[
%z_i^{(\ell,k)} = \max(z_{2i-1}^{(\ell-1,f)}, z_{2i}^{(\ell-1,f)}), \;\;\ell = 2,4,\cdots.
%\]
%The effects of pooling are two-fold: 1) it shrinks the size of the representation by half, thus quickly making up for the difference of sentence representation from the length variability, and 2) it helps filter out the undesired composition between words(see Section \ref{s:someAnalysis} for some analysis).
%

\subsection{Gating Network} \label{s:gate}
Previous CNNs, including those for NLP tasks~\cite{hu2014convolutional,KalchbrennerACL2014}, take a straightforward convolution-pooling strategy, in which the ``fusion" decisions (e.g., selecting the largest one in max-pooling) are made based on the values of feature-maps. This is essentially a soft template matching~\cite{cnnCV}, which works for tasks like classification, but undesired for maintaining the composition functionality of convolution. In this paper, we propose to use separate gating networks to release the scoring duty from the convolution, and let it focus on composition. Similar idea has been proposed by~\cite{SocherEtAl2011:RNN} for recursive neural networks on parsing tasks, but never been combined with a convolutional architecture.
\begin{figure}[h!]
\begin{center}
      \includegraphics[width=0.45\textwidth]{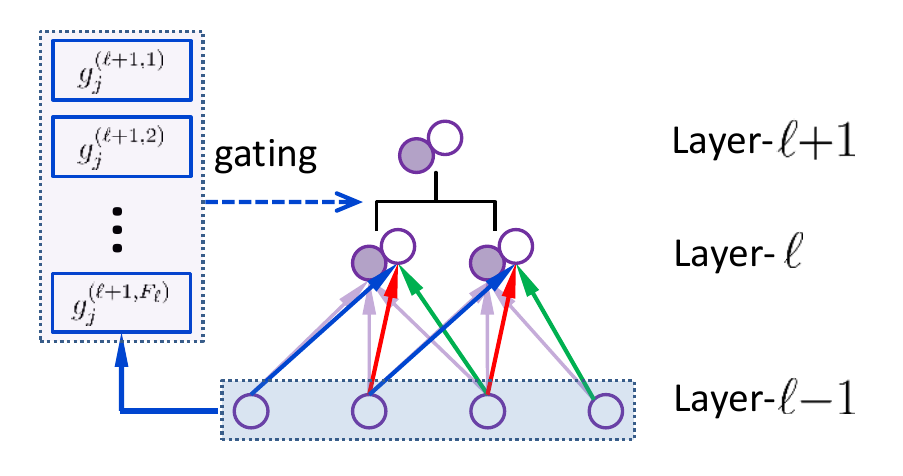}\\
    \caption{Illustration for gating network.}
    \label{f:gate}
  \end{center} %\vspace{-10pt}
\end{figure}

%More specifically, we use gating window with size $2$.
Suppose we have convolution feature-maps on Layer-$\ell$ and  gating (with window size = 2) on Layer-$\ell \hspace{-3pt} + \hspace{-3pt}1$. For the $j^{th}$ gating window  ($\small 2j\hspace{-3pt}-\hspace{-3pt}1, 2j$), we merge $\hat{\mathbf{z}}^{(\ell-1)}_{2j-1}$ and $\hat{\mathbf{z}}^{(\ell-1)}_{2j}$ as the input (denoted as  $\bar{\z}^{(\ell)}_j$) for gating network, as illustrated in Figure~\ref{f:gate}.
%\[
%\bar{\z}^{(\ell)}_j = [(\hat{\z}^{(\ell-1)}_{2j-1})^\top,\,\cdots,\, (\hat{\z}^{(\ell-1)}_{2j+k_{\ell}-1})^\top]^\top
%\]
%as the input, which merges the input to the $(2j\hspace{-2pt}-\hspace{-2pt}1)^{th}$ and $(2j)^{th}$ convolution window, as shown in Figure \ref{f:gate}.
We use a separate gate for each feature-map, but follow a different parametrization strategy for \textsc{Time-Flow} and \textsc{Time-Arrow}. With window size = 2, the gating is binary, we use a logistic regressor to determine the weights of two candidates. For $f\hspace{-3pt}\in \hspace{-3pt}\textsc{Time-Arrow}$, with location-dependent $\w^{(\ell,f, j)}_{\textsf{gate}}$, the normalized weight for \emph{left} side is
 \vspace{-5pt}
\begin{equation*}
g^{(\ell +1,f)}_j= 1/(1+e^{-\w^{(\ell,f,j)}_{\textsf{gate}}\bar{\z}^{(\ell)}_j }),  
 %p^{(\ell +1,f,j)}_{R}  = 1-p^{(\ell +1,f,j)}_{L}
%p^{(\ell+1, f,j)}_{R} &=& 1/(1+e^{\w^{(\ell,f)}_{g}\bar{\z}^{\ell}_j }),
\end{equation*}
while for For $f\hspace{-3pt}\in \hspace{-3pt}\textsc{Time-Flow}$, the parameters for the corresponding gating network, denoted as $\w^{(\ell,f)}_{\textsf{gate}}$, are shared.
%but follow a different parameter sharing strategy for \textsc{Time-Flow} and \textsc{Time-Arrow}:
%\begin{itemize}
%  \item For $f\hspace{-3pt}\in \hspace{-3pt}\textsc{Time-Flow}$, with the location-independent parameter $\w^{(\ell,f)}_{\textsf{gate}}$ the gating network assign the following weights to left window
%\begin{equation*}
%p^{(\ell +1,f,j)}_{L} = 1/(1+e^{-\w^{(\ell,f)}_{\textsf{gate}}\bar{\z}^{(\ell)}_j }), %p^{(\ell +1,f,j)}_{R}  = 1-p^{(\ell +1,f,j)}_{L}
%%p^{(\ell+1, f,j)}_{R} &=& 1/(1+e^{\w^{(\ell,f)}_{g}\bar{\z}^{\ell}_j }),
%\end{equation*}
%  \item For $f\hspace{-3pt}\in \hspace{-3pt}\textsc{Time-Arrow}$, with location-dependent parameter $\w^{(\ell,f, j)}_{\textsf{gate}}$, the weight to the left side is
% %\vspace{-5pt}
%\begin{equation*}
%p^{(\ell +1,f,j)}_{L} = 1/(1+e^{-\w^{(\ell,f,j)}_{\textsf{gate}}\bar{\z}^{(\ell)}_j }), %p^{(\ell +1,f,j)}_{R}  = 1-p^{(\ell +1,f,j)}_{L}
%%p^{(\ell+1, f,j)}_{R} &=& 1/(1+e^{\w^{(\ell,f)}_{g}\bar{\z}^{\ell}_j }),
%\end{equation*}
%\end{itemize}
The gated feature map is then a weighted sum to feature-maps from the two windows: 
{\small \begin{equation}\small
z^{(\ell+1,f)}_{j}= g^{(\ell +1,f)}_j z^{(\ell,f)}_{2j-1} + (1-g^{(\ell +1,f)}_j) z^{(\ell,f)}_{2j}.   \label{e:gate}
\end{equation}}
We find that this gating strategy works significantly better than direct pooling over feature-maps, and also slightly better than a hard gate version of Equation (\ref{e:gate}).

\subsection{Recursive Architecture} \label{s:recursive}
\begin{figure}[h!]
\begin{center}
 \includegraphics[width=0.65\textwidth]{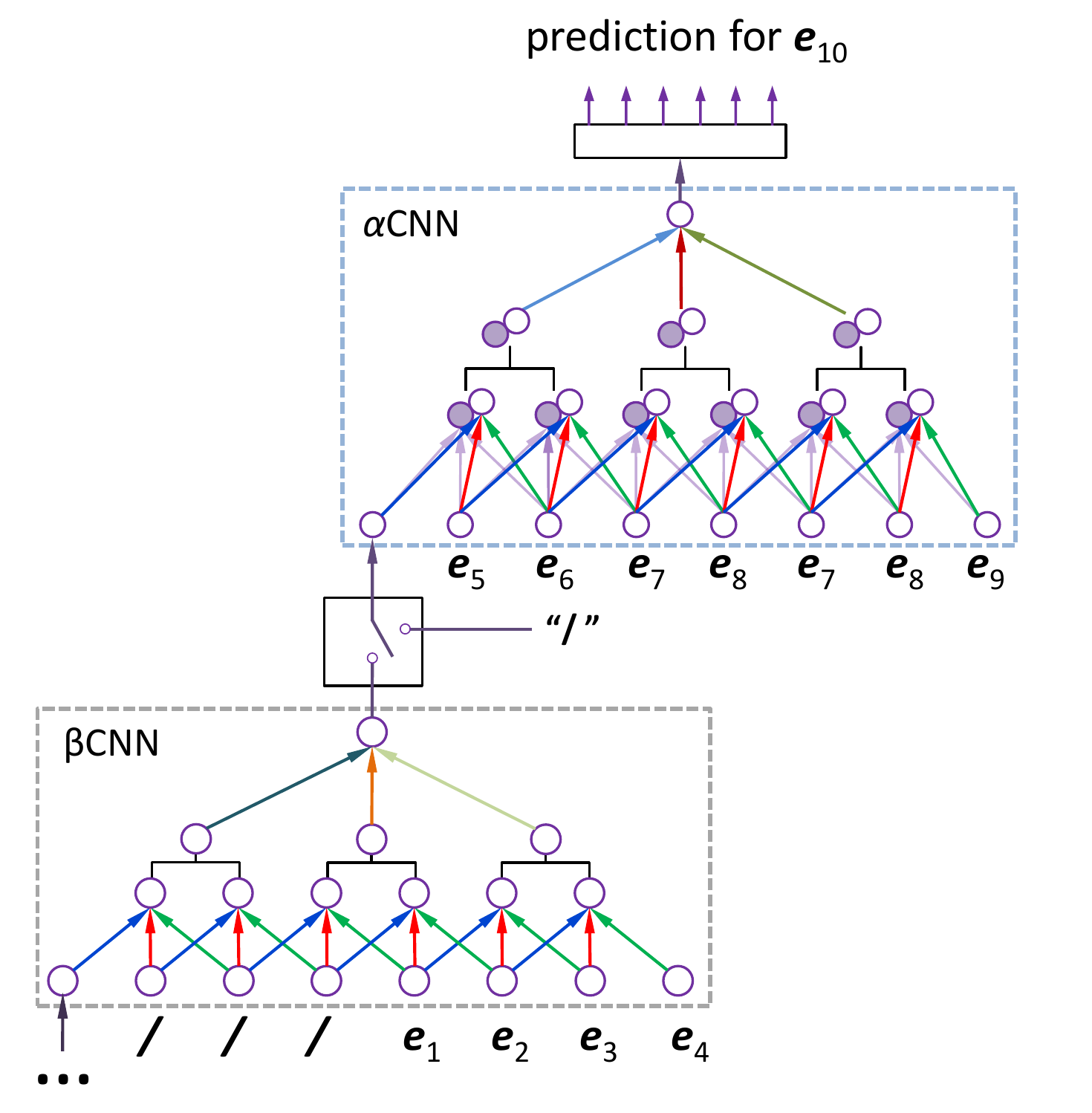}  
          \caption{$gen$CNN with recursive structure.}
    \label{f:recursive}
  \end{center} 
\end{figure}

As suggested early on in Section~\ref{s:overview} and Figure~\ref{f:overall}, we use extra CNNs with conventional weight-sharing, named $\beta$CNN, to summarize the history out of scope of $\alpha$CNN. More specifically, the output of $\beta$CNN (with the same dimension of word-embedding) is put before the first word as the input to the $\alpha$CNN, as illustrated in Figure~\ref{f:recursive}. Different from $\alpha$CNN, $\beta$CNN is designed just to summarize the history, with weight shared across its convolution units. In a sense, $\beta$CNN has only \textsc{Time-Flow} feature-maps. All $\beta$CNN are identical and recursively aligned, enabling $gen$CNN to handle sentences with arbitrary length. We put a special switch after each $\beta$CNN to turn it off (replacing a pading vector shown as ``$/$" in Figure~\ref{f:recursive}) when there is no history assigned to it. As the result, when the history is shorter than $L_\alpha$, the recursive structure reduces to $\alpha$CNN.

In practice, $90$+\% sentences can be modeled by $\alpha$CNN with $L_\alpha = 40$ and  $99$+\% sentences can be contained with one extra $\beta$CNN. Our experiment shows that this recursive strategy yields better estimate of conditional density than neglecting the out-of-scope history (Section \ref{s:fbis}).  In practice, we found that a larger (greater $L_\alpha$) and deeper $\alpha$CNN works better than small $\alpha$CNN and more recursion of $\beta$CNN, which is consistent with our intuition that the bottom-up convolutional architecture is well suited for modeling the sequence.

\subsection{Analysis} \label{s:analysis}
%We now give a more detailed analysis on the key components of $gen$CNN and the mechanism of their interaction.

\subsubsection{\textsc{Time-Flow} vs. \textsc{Time-Arrow}}
Both conceptually and systemically, $gen$CNN gives two interweaved treatments of word history. With the globally-shared parameters in the convolution units, \textsc{Time-Flow} summarizes \emph{what has been said}. The hierarchical convolution+gating architecture in \textsc{Time-Flow} enables it to model the composition in language, yielding representation of segments at different intermediate layers. \textsc{Time-Flow} is aware of the sequential direction, inherited from the space-awareness of CNN, but it is not sensitive enough about the prediction task, due to the uniform weights in the convolution.

On the other hand, \textsc{Time-Arrow}, living in location-dependent parameters of convolution units,  acts like an arrow pin-pointing the prediction task.   \textsc{Time-Arrow} has predictive power all by itself, but it concentrates on capturing the direction of time and consequently short on modelling the long-range dependency.

\textsc{Time-Flow} and \textsc{Time-Arrow} have to work together for optimal performance in predicting \emph{what is going to be said}. This intuition has been empirically verified, as our experiments have demonstrated that \textsc{Time-Flow} or \textsc{Time-Arrow} alone perform inferiorly. One can imagine, through the layer-by-layer convolution and gating, the \textsc{Time-Arrow} gradually picks the most relevant part from the representation of \textsc{Time-Flow} for the prediction task, even if that part is long distance  ahead.

%\vspace{-5pt}
%\begin{itemize}
%  \item {\bf \textsc{Time-Flow}:} with the globally-shared parameters in the convolution units, \textsc{Time-Flow} summarizes, in different intermediate layers, \emph{what has been said}. The convolution+gating architecture in \textsc{Time-Flow} enables it to well model the composition in language, yielding representation of segments at different resolutions. \textsc{Time-Flow} is aware of the sequential direction, inherited from the space-awareness of CNN, but it is not sensitive enough about the prediction task, due to the uniform weights in the convolution. \vspace{-5pt}
%  \item {\bf \textsc{Time-Arrow}:} living in the individual parameters of convolution units, it acts like an arrow pin-pointing to the prediction task.   \textsc{Time-Arrow} has predictive power all by itself, just like a conventional language model, but it concentrates on capturing the direction of time and therefore short at modelling the long-range dependency.
%\end{itemize}

\subsubsection{$gen$CNN vs. RNN-LM}
Different from RNNs, which recursively applies a relatively simple processing units, $gen$CNN gains its ability on sequence modeling mostly from its flexible and powerful bottom-up and convolution architecture. $gen$CNN takes the ``uncompressed" history, therefore avoids
\begin{itemize}
  \item  the difficulty in finding the representation for history (i.e., unfinished sentences), especially those end in the middle of a chunk (e.g.,``\texttt{\small the cat sat on the}"),
  \item the damping effort in RNN when the history-summarizing hidden states are updated at each time, which renders the long term memory rather difficult.
\end{itemize}
Both drawbacks can only be partially ameliorated with complicated design of gates~\cite{oriLSTM} and or more heavy processing units (essentially a fully connected DNN)~\cite{sutskever2014sequence}.
\section{$gen$CNN: Training} \label{s:learning}
The parameters of a $gen$CNN $\bar{\Theta}$ consists of the parameters for CNN $\Theta_{nn}$,  word-embedding $\Theta_{embed}$, and the parameters for soft-max $\Theta_{softmax}$. All the parameters are jointly learned by maximizing the likelihood of observed sentences.
Formally the log-likelihood of sentence $\calS_n$ ($\,\overset{\text{def}}{=}[\e^{(n)}_1,\e^{(n)}_2,\cdots,\e^{(n)}_{T_n}]$)  is 
\[
\log p(\calS_n; \bar{\Theta}) = \sum_{t=1}^{T_n} \log p(\e^{(n)}_t| \e^{(n)}_{1:t-1};  \bar{\Theta}),
\]
which can be trivially split into $T_n$ training instances during the optimization, in contrast to the training of RNN that requires unfolding through time due to the temporal-dependency of the hidden states.
%and log-likelihood of a training corpus $\calC$ is  \vspace{-5pt}
%\[
%\log p(\calC; \bar{\Theta}) = \sum_{\calS_n \in \calC} \log p(\calS_n; \bar{\Theta}).
%\]

\subsection{Implementation Details} \label{s:details}
%In this section, we give implementation details.
\paragraph{Architectures:} In all of our experiments (Section~\ref{s:expts_generation} and \ref{s:expts_LM})£¬ we set the maximum words for $\alpha$CNN to be 30 and that for $\beta$CNN to be 20.  $\alpha$CNN have two convolution layers (both containing \textsc{Time-Flow} and \textsc{Time-Arrow} convolution) and two gating layers, followed by a fully connected layer (400 dimension)  and then a soft-max layer. The numbers of feature-maps for \textsc{Time-Flow} are respectively 150 (1st convolution layer) and 100 (2nd convolution layer), while \textsc{Time-Arrow} has the same feature-maps. $\beta$CNN is relatively simple, with two convolution layer containing only \textsc{Time-Flow} with 150 feature-maps, two gating layers and a fully connected layer. We use ReLU as the activation function for convolution layers and switch to Sigmoid for fully connected layers.
We use word embedding with dimension 100.

\paragraph{Soft-max:} Calculating a full soft-max is expensive since it has to enumerate all the words in vocabulary (in our case 40K words) in the denominator. Here we take a simple hierarchical approximation of it, following~\cite{bahdanau2014neural}. Basically we group the words into 200 clusters (indexed by $c_m$), and factorize (in an approximate sense) the conditional probability of a word $p(\e_t|\e_{1:t-1}; \bar{\Theta})$ into the probability of its cluster and the probability of $\e_t$ given its cluster 
\[
p(c_m|\e_{1:t-1}; \bar{\Theta})\,p(\e_t|c_m; \Theta_{softmax}). 
\]
We found that this simple heuristic can speed-up the optimization by 5 times with only slight loss of accuracy.

\paragraph{Optimization:} We use stochastic gradient descent with mini-batch (size 500) for optimization, aided further by AdaGrad~\cite{adagrad}. For initialization, we use Word2Vec~\cite{word2vec} for the starting state of the word-embeddings (trained on the same dataset as the main task), and set all the other parameters by randomly sampling from uniform distribution in $[-0.1,0.1]$. The optimization is done mainly on a Tesla K40 GPU, which takes about 2 days for the training on a dataset containing 1M sentences.

%\begin{itemize}
%  \item h-softmax
%  \item adgrad We use standart sgd and perform adagrade \cite{adagrade} to optimise.
%  \item initialization
%  follow \cite{wordembeding} the word embeding matrix is pre-trained with DNN.  In this
%  work, we employ the toolkit Word2Vec, to pre-train the word embedding on the training data. All the other weights an random initialize between -0.1 and 0,1.
%  \item training time we perform our experiments on a single Tesla K40 GPU. The training set contains about 27M  words. The embeding size is set to 100,
%  the window is set to 30. the vocabulary size is 40000. It takes about 2 days to finish our training.
%  \item parallelization
%\end{itemize}%\vspace{50pt} %\newpage
\section{Experiments: Sentence Generation} \label{s:expts_generation}

In this experiment, we randomly generate sentences by recurrently  sampling  
\[
\e_{t+1}^\star \sim p(\e_{t+1}|\e_{1:t}; \bar{\Theta}), 
\]
and put the newly generated word into history,
until \texttt{\small EOS} (end-of-sentence) is generated.  We consider generating two types of sentences: 1) the plain sentences, and 2) sentences with dependency parsing, which will be covered respectively in Section~\ref{s:genNLS} and \ref{s:genDTree}.

\begin{table*}[t!]
	\begin{center}
		\begin{tabular}{|l|}

\hline  \texttt{\small  \underline{\color{blue}``} we are in the building of china 's social development and the businessmen} \\ \texttt{\small audience , '' he said .}\\
\hline \texttt{\small  \underline{\color{blue}clinton} was born in DDDD , and was educated at the university of edinburgh.} \\

\hline \texttt{\small  \underline{\color{blue}bush} 's first album , `` the man '' , was released on DD november DDDD .} \\

\hline \texttt{\small \underline{\color{blue}it is one} of the first section of the act in which one is covered in real }  \\
 \texttt{\small place that recorded in norway .}\\

\hline\hline

%jiang
%zemin pointed out the democratic people’s republic of korea ( dprk ) , by vigorously implementing the ban on a missile move in south korea , added kim recently quoted by a chinese saying that science and technology in international cooperation was not affected .

\texttt{\small  this objective is brought to us the welfare of our country } \\
\hline
\texttt{\small
russian president putin delivered a speech to the sponsored by the 15th asia
} \\
\texttt{\small
pacific economic cooperation ( apec ) meeting in an historical arena on oct .
}\\

\hline \texttt{\small
light and snow came in kuwait and became operational , but was rarely
} \\
\texttt{\small placed in houston . } \\

\hline \texttt{\small johnson became a drama company in the DDDDs , a television broadcasting } \\
\texttt{\small  company owned by the broadcasting program .}\\

\hline
\hline

 \texttt{\small  ( ( the two $\star$ sides ) $\star$ should ( $\star$ assume ( a strong $\star$ target ) ) ) . )
}\\

\hline  \texttt{\small  ( it $\star$ is time ( $\star$ in ( every $\star$ country ) $\star$ signed ( the $\star$ speech ) ) . )
}\\

\hline  \texttt{\small
( ( initial $\star$ investigations ) $\star$ showed ( $\star$ that ( spot $\star$ could ( $\star$ be (
}\\
\texttt{\small
further $\star$ improved significantly  ) ) . )
}\\

\hline \texttt{\small  ( ( a  $\star$ book ( to $\star$ northern ( the 21 st $\star$ century ) ) ) . )} \\
\hline

		\end{tabular}
	\end{center}
	\caption{\label{t:genExamples} Examples of sentences generated by $gen$CNN. In the upper block (row 1-4) the underline words are given by the human; In the middle block (row 5-8), all the sentences are generated without any hint. The bottom block (row 9-12) shows the sentences with dependency tag generated by $gen$CNN trained with parsed examples.}
\end{table*}

\subsection{Natural Sentences} \label{s:genNLS}
We train $gen$CNN on Wiki data with 112M words  for one week, with some representative examples randomly generated given in
%with xxx, and digits replaced with ``\texttt{\small DDDD}"
Table \ref{t:genExamples} (upper and middle blocks). We try two settings, by asking $gen$CNN to 1) finish a sentence started by human (upper block), or 2) generate a sentence from the beginning (middle block), or   It is fairly clear that most of the time  $gen$CNN can generate sentences that are syntactically grammatical and semantically meaningful. More specifically, most of the sentences can be aligned to a parse tree with reasonable structure. It is also worth noting that quotation marks (\texttt{\small ``} and \texttt{\small ''}) are always generated in pairs and in the correct order, even across a relatively long distance, as exemplified by the first sentence in the upper block.

%\begin{itemize}
%  \item \texttt{\small  \underline{\color{blue}clinton} was born in DDDD , and was educated at the university of edinburgh} \vspace{-5pt}
%  \item \texttt{\small  \underline{\color{blue}bush} 's first album , `` the man '' , was released on DD november DDDD .} \vspace{-5pt}
%  \item \texttt{\small  \underline{\color{blue}``} we are in the building of china 's social development and the businessmen
%  audience , '' he said .}
%\end{itemize}

\subsection{Sentences with Dependency Tags} \label{s:genDTree} 
For training, we first parse\cite{depparser} the English sentences and feed sequences with dependency tags as follows 
\[
\texttt{\small ( I $\star$ like ( red $\star$ apple ) )} 
\]
to $gen$CNN, where 1) each paired parentheses contain a subtree, and 2) the symbol ``$\star$" indicates that the word next to it is the dependency head in the corresponding sub-tree. Some representative examples generated by $gen$CNN are given in Table~\ref{t:genExamples} (bottom block). As it suggests, $gen$CNN is fairly accurate on respecting the rules of parentheses, and probably more remarkably, it can get the dependency tree head correct most of the time.

\section{Experiments: Language Modeling} \label{s:expts_LM}
We evaluate our model as a language model in terms of both perplexity~\cite{ppl} and its efficacy in re-ranking the $n$-best candidates from state-of-the-art models in statistical machine translation, both with comparison to the following competitor language models.

\paragraph{Competitor Models} we compare $gen$CNN to the following competitor models
\begin{itemize}
  \item 5-gram: { We use SRI Language Modeling Toolkit~\cite{stolcke2002srilm} to train a 5-gram language model with modified Kneser-Ney smoothing}; 
  \item FFN-LM: The neural language model based on feedfoward network~\cite{nplm}. We vary the input window-size from 5 to 20, while the performance stops improving after window size 20; 
  \item RNN: we use the implementation\footnote{http://rnnlm.org/} of RNN-based language model with hidden size 600 for optimal performance of it;
  \item LSTM: we use the code in Groundhog\footnote{https://github.com/lisa-groundhog/GroundHog}, but vary the hyper-parameters, including the depth and word-embedding dimension, for best performance. LSTM~\cite{oriLSTM} is widely considered to be the state-of-the-art for sequence modeling. 
\end{itemize}

\subsection{Perplexity}
%Two experiments are carried out, with the second one designed to test $gen$CNN's ability on modeling long range correlation.
We test the performance of $gen$CNN on  \textsc{Penn Treebank} and \textsc{FBIS}, two public datasets with different sizes.

\subsubsection{On \textsc{Penn Treebank}} \label{s:penn}
Although a relatively small dataset \footnote{http://www.fit.vutbr.cz/$\sim$imikolov/rnnlm/simple-examples.tgz},  \textsc{Penn Treebank} is widely used as a language modelling benchmark ~\cite{SequenceGen_Graves13,rnnMikolov}. It has $930,000$ words in training set, $74,000$ words in validation set, and  $82,000$ words in test set.
We use exactly the same settings as in~\cite{rnnMikolov}, with a  $10,000$-words vocabulary (all out-of-vocabulary words are replaced with \texttt{unknown}) and end-of-sentence token (\texttt{EOS}).
%While using exactly same settings (The vocabulary is limited to 10,000 words, with all other words mapped to a special ??unknown word? ¡¥ token. The end-of -sentence token was included in the input sequences, and was counted in the sequence loss.) This allow us to compare directly with ours.
In addition to the conventional testing strategy where the models are kept unchanged during testing, \newcite{rnnMikolov} proposes to also update the parameters in an online fashion when seeing test sentences. This new way of testing, named ``dynamic evaluation", is also adopted by \newcite{SequenceGen_Graves13}.

From Table~\ref{t:penn},  $gen$CNN manages to give perplexity superior in both metrics, with about $25$ point reduction over the widely used 5-gram, and over $10$ point reduction from LSTM, the state-of-the-art and the second-best performer.
We defer the comparison of $gen$CNN variants to next experiment on a larger dataset (\textsc{FBIS}), since \textsc{Penn Treebank} is too small for evaluating some of the differences between them.

%On this small data set, $\alpha$CNN yields almost identical result as $gen$CNN (result omitted).

%Neural networks are usually evaluated on test data with fixed weights. For
%prediction problems however, where the inputs are the targets, it is legitimate
%to allow the network to adapt its weights as it is being evaluated (so long as
%it only sees the test data once). Mikolov refers to this as dynamic evaluation.
%Dynamic evaluation allows for a fairer comparison with compression algorithms,
%for which there is no division between training and test sets, as all data is only predicted once.

\begin{table}[h]
	\begin{center}
		\begin{tabular}{|l|c|c|}
			\hline \bf Model &  \bf Perplexity & \bf Dynamic \\ \hline
			5-gram, KN5 & 141.2 & --\\
			\hline
			FFNN-LM & 140.2 & -- \\
			\hline
			RNN & 124.7 & 123.2 \\
			\hline
			LSTM & 126 &117 \\
			\hline
			$gen$CNN & \textbf{116.4} & \textbf{106.3} \\
			\hline
%			\hline
%			RNNLM (dynamic) & 123.2 \\ \hline
%			LSTMLM (dynamic) & 117 \\ \hline
%			$gen$CNN (dynamic) & \textbf{106.3} \\
%			\hline
		\end{tabular}
	\end{center} 
	\caption{\label{t:penn} \textsc{Penn Treebank} results, where the 3rd column are the perplexity in dynamic evaluation, while the numbers for RNN and LSTM are taken as reported in the paper cited above. The numbers in boldface indicate that the result is significantly better than \emph{all competitors} in the same setting.}
\end{table}

\subsubsection{On \textsc{FBIS}} \label{s:fbis}
%We run the second experiment on a relative large dataset.
The FBIS corpus (LDC2003E14) is relatively large, with 22.5K sentences and 8.6M English words. The validation set is NIST MT06 and test set is NIST MT08.
For training the neural network, we limit the vocabulary to the most frequent 40,000 words, covering $\sim\hspace{-2pt} 99.4 \%$ of the corpus. Similar to the first experiment, all out-of-vocabulary words are replaced with \texttt{unknown} and the \texttt{EOS} token  is counted in the sequence loss.
\begin{table}[h!]
	\begin{center}
		\begin{tabular}{|l|r|}
			\hline \bf Model & \bf Perplexity \\ \hline
			5-gram, KN5 & 278.6 \\ \hline
			FFN-LM(5-gram) & 248.3 \\ \hline
			FFN-LM(20-gram) & 228.2 \\ \hline
			RNN & 223.4 \\ \hline
			LSTM & 206.9 \\ \hline
			 $gen$CNN & \textbf{181.2} \\

			\hline \hline
 			\textsc{Time-Arrow} only  & 192 \\
 			\hline
 			\textsc{Time-Flow} only & 203 \\
 			\hline
 			 $\alpha$CNN only & 184.4 \\
           \hline
 		\end{tabular}
	\end{center}
	\caption{\label{t:fbis} \textsc{FBIS} results. The upper block (row 1-6) compares $gen$CNN and the competitor models, and the bottom block (row 7-9) compares different variants of $gen$CNN. }
\end{table}

From Table~\ref{t:fbis} (upper block), $gen$CNN clearly wins again in the comparison to competitors, with over 25 point margin over LSTM (in its optimal setting), the second best performer. Interestingly $gen$CNN outperforms its variants also quite significantly (bottom block): 1) with only \textsc{Time-Arrow} (same number of feature-maps), the performance deteriorates considerably for losing the ability of capturing long range correlation reliably; 2) with only \textsc{Time-Flow} the performance gets even worse, for partially losing the sensitivity to the prediction task. It is quite remarkable that, although $\alpha$CNN (with $L_\alpha=30$) can achieve good results, the recursive structure in full $gen$CNN can further decrease the perplexity by over 3 points, indicating that $gen$CNN can benefit from modeling the dependency over range as long as 30 words.

%\subsubsection{Correlation in Distance}
%In stead of just evaluating the perplexity on entire sentence, we also devise an experiment to test $gen$CNN's ability in modeling the long range dependency.
%%The experiment is designed to study how the correlation decays with the word distance.
%More specifically, we evaluate how the conditional probability of a word given by $gen$CNN changes after randomly replacing the word $k$ places before it in a sentence, and see how the influence decays with $k$. From Figure~\ref{f:longrange}, the effect of a proceeding word in $gen$CNN, although deceases with distance, does not damp to zero even after xx words, suggesting that $gen$CNN does maintain a substantial long range dependency.
%
%\begin{figure}[h!]
%\begin{center}
% \includegraphics[width=0.35\textwidth]{pics/pooling.png}  \vspace{-20pt}
%          \caption{long range dependency  [space holder].}
%    \label{f:longrange}
%  \end{center}
%\end{figure}
%

\subsection{Re-ranking for Machine Translation}
In this experiment, we re-rank the $1000$-best English translation candidates for Chinese sentences generated by statistical machine translation (SMT) system, and compare it with other language models in the same setting. 

%based on 1) merely the perplexity from $gen$CNN, 2) a scoring function with perplexity of $gen$CNN as an important feature. As shown in  Table \ref{t:rerank}, re-ranking with just $gen$CNN perplexity can yield $1.2$ and $1.4$ increase of BLEU score, a vast improvement for a single feature.

\paragraph{SMT setup}
The baseline hierarchical phrase-based SMT system ( Chines$\rightarrow$ English) was
built using Moses, a widely accepted state-of-the-art, with default settings. The bilingual training data is from NIST MT2012 constrained track, with reduced size of 1.1M sentence pairs using selection strategy in~\cite{Axelrod}. The baseline use conventional 5-gram language model (LM), estimated with
modified Kneser-Ney smoothing~\cite{ChenGoodman} on the English side of the 329M-word Xinhua portion of English Gigaword(LDC2011T07). We also try FFN-LM, as a much stronger language model in decoding. The weights of all the features are tuned via MERT~\cite{och2002discriminative} on NIST MT05, and tested on NIST MT06 and MT08. Case-insensitive NIST BLEU\footnote{ftp://jaguar.ncsl.nist.gov/mt/resources/mteval-v11b.pl} is used in evaluation.

Re-ranking with $gen$CNN significantly improves the quality of the final translation. Indeed, it can increase the BLEU score by over 1.33 point over Moses baseline on average. This boosting force barely slacks up on translation with a enhanced language model in decoding: $gen$CNN re-ranker still achieves 1.29 point improvement on top of Moses with FFN-LM, which is 1.76 point over the Moses (default setting). To see the significance of this improvement, the state-of-the-art Neural Network Joint Model~\cite{devlin2014} usually brings less than one point increase on this task.

\begin{table}[t!]
	\begin{center}
		\begin{tabular}{|l|c|c|c|}
			\hline \bf Models &  \bf MT06 & \bf MT08 & \bf Ave. \\ \hline
			Baseline & 38.63 & 31.11 & 34.87 \\
			\hline
            %re-ranking by & & &\\
			\hline
            RNN rerank & 39.03 & 31.50 & 35.26 \\
			\hline
			LSTM rerank & 39.20 & 31.90 & 35.55  \\
			\hline
			FFN-LM rerank & 38.93 & 31.41 & 35.14 \\
			\hline
			$gen$CNN rerank& \textbf{39.90} & \textbf{32.50} & \textbf{36.20} \\
			\hline
			\hline
			Base+FFN-LM & 39.08 & 31.60 & 35.34 \\
			\hline
			$gen$CNN rerank & \textbf{40.4} & \textbf{32.85} & \textbf{36.63} \\
			\hline
%			RNNLM (dynamic) & 123.2 \\ \hline
%			LSTMLM (dynamic) & 117 \\ \hline
%			$gen$CNN (dynamic) & \textbf{106.3} \\
%			\hline
		\end{tabular}
	\end{center} 
	\caption{\label{rank table} The results for re-ranking the 1000-best of Moses. Note that the two bottom rows are on a baseline with enhanced LM.}
\end{table}

\section{Related Work} \label{s:related} 
In addition to the long thread of work on neural network based language model~\cite{auli2013,rnnMikolov,SequenceGen_Graves13,Bengio03aneural,nplm}, our work is also related to the effort on modeling long range dependency in word sequence prediction\cite{wu2003maximum}. Different from those work on hand-crafting features for incorporating long range dependency, our model can elegantly assimilate relevant information in an unified way, in both long and short range, with the bottom-up information flow and convolutional architecture.

%Syntax-based language model

%{\color{blue} [Maybe we don't need this section at all]}
%\begin{itemize}
%  \item RNN, LSTM: maybe say that in the analysis section?
%  \item RNN with tensor structure?
%  \item FNNLM: fully connected layers, with performance inferior to recurrent neural network
%  \item RAE
%  \item Wu Jun's work?
%\end{itemize}

CNN has been widely used in computer vision and speech ~\cite{cnnCV,krizhevsky2012imagenet,lecun1995convolutional,abdel2012applying}, and lately in sentence representation\cite{kalchbrenner2013}, matching\cite{hu2014convolutional} and classification\cite{KalchbrennerACL2014}. To our best knowledge, it is the first time this is used in word sequence prediction. Model-wise the previous work that is closest to $gen$CNN is the convolution model for predicting moves in the Go game~\cite{go}, which, when applied recurrently, essentially generates a sequence. Different from the conventional CNN taken in~\cite{go}, $gen$CNN has architectures designed for modeling the composition in natural language and the temporal structure of word sequence.

\section{Conclusion} 
We propose a convolutional architecture for natural language generation and modeling.  Our extensive experiments on sentence generation, perplexity, and $n$-best re-ranking for machine translation show that our model can significantly improve upon state-of-the-arts.

\newpage
\bibliographystyle{acl}
\bibliography{acl2015}
\end{document}